\documentclass[10pt,twocolumn,letterpaper]{article}
\usepackage{cvpr}
\usepackage{times}
\usepackage{epsfig}
\usepackage{graphicx}
\usepackage{amsmath}
\usepackage{amssymb}
\usepackage{multirow}
\usepackage{cite}

\usepackage[pagebackref=true,breaklinks=true,letterpaper=true,colorlinks,bookmarks=false]{hyperref}

\cvprfinalcopy 

\begin{document}

\title{Integrating Coarse Granularity Part-level Features with Supervised Global-level Features for Person Re-identification}

\author{Xiaofei MAO\textsuperscript{1}, Jiahao Cao\textsuperscript{2}, Dongfang Li\textsuperscript{3}, Xia Jia\textsuperscript{*}, Qingfang Zheng\textsuperscript{*}\\
ZTE Corporation\\
{\tt\small xfmao99@gmail.com {cao.jiahao, li.dongfang}@zte.com.cn}
}

\maketitle

\begin{abstract}
   Person re-identification (Re-ID) has achieved great progress in recent years. However, person Re-ID methods are still suffering from body part missing and occlusion, where the learned representations are less reliable. In this paper, we propose a robust coarse granularity part-level person Re-ID network (CGPN), which extracts robust regional features and integrates supervised global features for pedestrian images. CGPN gains two-fold benefit toward higher accuracy for person Re-ID. On one hand, CGPN learns to extract effective regional features for pedestrian images. On the other hand, compared with extracting global features directly by backbone network, CGPN learns to extract more accurate global features with a supervision strategy. The single model trained on three Re-ID datasets achieves state-of-the-art performances and outperforms any existing approaches. Especially on CUHK03, which is the most challenging Re-ID dataset, we obtain a top result of Rank-1/mAP=87.1\%/83.6\% without re-ranking, outperforming the current best method by +7.0\%/+6.7\%.
\end{abstract}

\section{Introduction}

\begin{figure}[t]
\begin{center}
\includegraphics[width=0.8\linewidth]{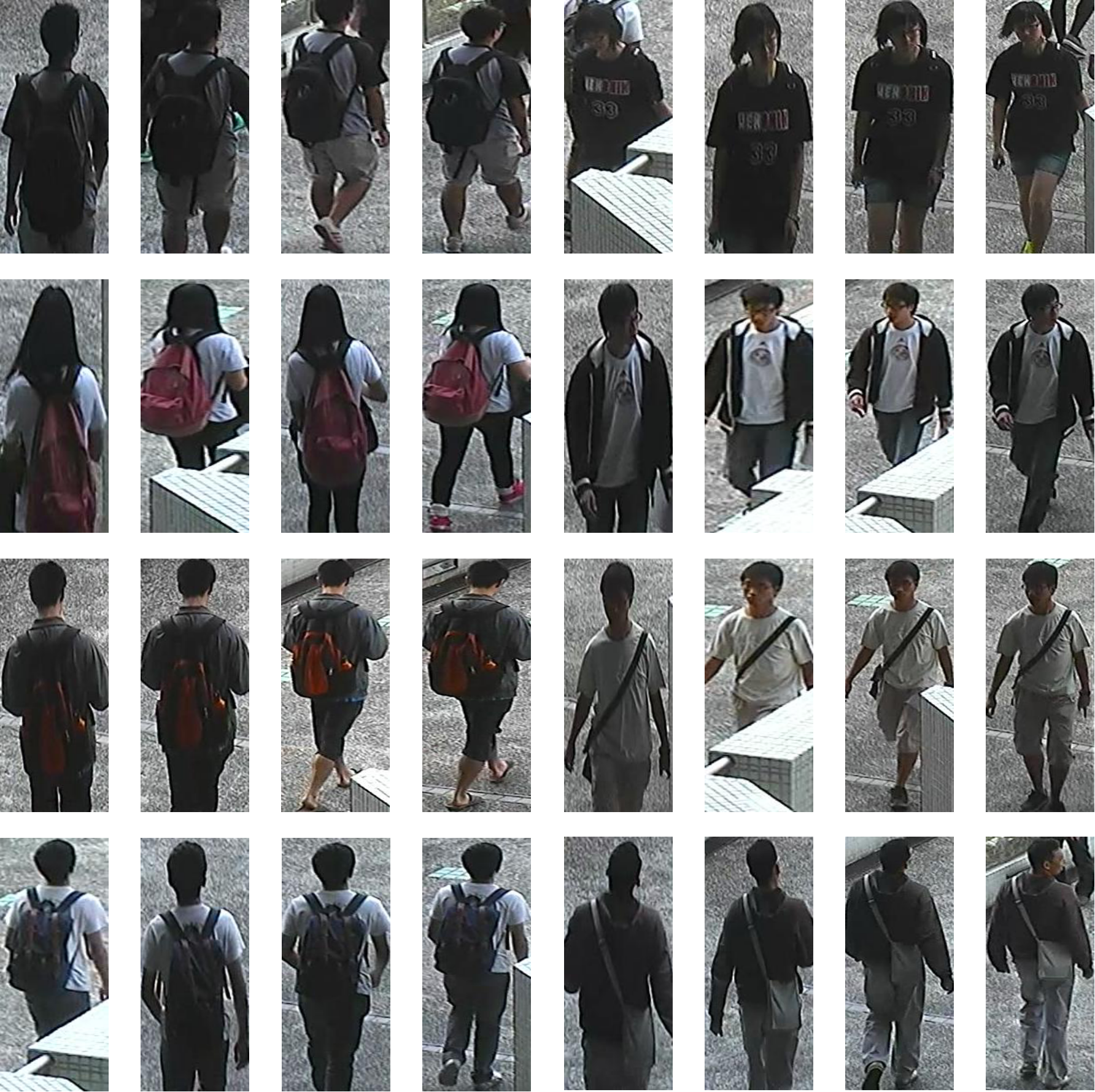}
\end{center}
	\caption{Pedestrian images in CUHK03-labeled dataset. We can see that for the same person in realistic scenarios, occluded pedestrian images and body-part missed pedestrian images are both captured as people moving around the cameras.}
\label{fig:cuhk03_data}
\end{figure}

Person re-identification (Re-ID) aims to retrieve a given person among all the gallery pedestrian images captured across different cameras. It is a challenging task to learn robust pedestrian feature representations as realistic scenarios are highly complicated with regards to illumination, background, and occlusions. In recent years, person Re-ID has achieved great progress \cite{recentprogresschang2018multi,recentprogressliu2017end,recentprogresssaquib2018pose,recentprogressshen2018deep,mgn,pcb,pyramid}. However, person Re-ID methods are still suffering from occluded or body part missed pedestrian images, where they fail to extract discriminative deep features for person Re-ID. Intuitively, the complexity of realistic scenarios increases the difficulty to make correct retrieval for person Re-ID \cite{misalignmentshen2015person,misalignmentvarior2016siamese,PBF}. Therefore, the performance of existing person Re-ID methods usually decreases a lot when dealing with realistic person Re-ID dataset like CUHK03 which contains a lot of occluded or body part missed pedestrian images, as illustrated in Fig. \ref{fig:cuhk03_data}.

As it is well-known, part-based methods \cite{mgn,pcb,pyramid} like MGN\cite{mgn} are widely used in person Re-ID and achieve promising performance. Generally, part-based methods learn to combine global features and discriminative regional features for person Re-ID. The global features
in part-based methods are usually extracted directly from the whole person image by the backbone network, while the regional features are generated by directly partitioning feature maps of the whole body into a fixed number of parts. Nevertheless, the overall performance of such part-based methods seriously depends on that all person images are well-bounded holistic person images with
little occlusions or body part missing. As real-world scenarios are complicated, the bounding boxes detected by the detection algorithm may not be accurate enough, which usually leads to occluded or body part missed pedestrian images as Fig. \ref{fig:cuhk03_data} shows. When dealing with such occluded or body part missed pedestrian images, the global features extracted from the whole image directly by the backbone network become less accurate, moreover, the regional features generated by directly partitioning feature maps of the whole body may focus on occluded parts and become ineffective, which impair the person Re-ID accuracy evidently.

To address above problems, in this paper, we propose the CGPN model that learns discriminative and diverse feature representations without using any third models. Our CGPN model can be trained end to end and performs well on three person Re-ID datasets. Especially on CUHK03, which contains a lot of occluded or body part missed pedestrian images, our method achieves state-of-the-art performances and outperforms the current best method by a large margin. CGPN
has three branches, each branch consists of a global part and a local part. The global part is supervised to learn more accurate global features by part-level body regions. With the supervision strategy, the global part can learn more proper global features for occluded or body part missed pedestrian images. For the local part, as pedestrian images detected in realistic scenarios are often
occluded or body-part missed, too much finer grained local features generated by partitioning the whole body feature maps may decrease model performance, we propose a coarse grained part-level feature strategy which can extract effective regional features and performs better on the three person Re-ID datasets.

CGPN gains two-fold benefit toward higher accuracy for person Re-ID. Firstly, compared with extracting global features directly by backbone network, CGPN learns to extract more accurate global features with the supervision strategy. Secondly, with the coarse grained part-level feature strategy, CGPN is capable to extract effective body part features as regional features for person
Re-ID. Besides, our method is completely an end-to-end learning process, which is easy for learning and implementation. Experimental results confirm that our method achieves state-of-the-art performances on several mainstream Re-ID datasets, especially on CUHK03, the most challenging dataset for person Re-ID, in single query mode, we obtain a top result of Rank-1/mAP=87.1\%/83.6\% with this method without re-ranking, outperforming the current best
method by +7.0\%/+6.7\%.

The main contributions of our work are summarized as follows:
\begin{itemize}
	\item We propose a novel framework named CGPN, which effectively integrates coarse grained part-level features and supervised global-level features and is more robust for person Re-ID.
	\item We firstly develop the coarse grained part-level feature strategy for person Re-ID.
	\item We prove the integration model of coarse grained part-level features and supervised global-level features achieves state-of-the-art results on three Re-ID datasets, especially on CUHK03 dataset, our model outperforms the current best method by a large margin.
\end{itemize}

\section{Related works}
\subsection{Part-based Re-ID model}

As deep learning is widely used in person Re-ID nowadays, most existing methods
\cite{li2014deepreid, deepmetric} choose to extract feature maps by directly applying a deep convolution network such as ResNet \cite{he2016deep}. However, single global feature extracted from the whole person image by deep convolution network not performs well as expected. The reason is that person images captured by cameras usually contain random background information and are often occluded or body part missed which impair the performance a lot. Then part-based methods are proposed to try to get additional useful local information from person images for person Re-ID. As an effective way to extract local features, part-based methods \cite{pcb, mgn, pyramid, hacnn} usually benefit from person structure and together with global features, push the performance of person Re-ID to a new level. The common solution of part-based methods is to split the feature maps horizontally into several parts according to human body structure and concatenate the feature maps of each part. However, when dealing with
occluded or body part missed pedestrian images we find that part-based methods like MGN \cite{mgn} which achieved state-of-the-art results on person Re-ID datasets meet performance decrease. Obviously, part-based methods are common solutions to holistic pedestrian images as they can get correct body parts by uniform partitioning, however, these methods are less effective to occluded or body part missed pedestrian images.

\subsection{Attention-based Re-ID model}
Recently, some attention-based methods try to address the occlusion or body-part missing problems with the help of attention mechanism. Attention module is developed to help to extract more accurate features by locating the significant body parts and learning the discriminative features from these informative regions. Li et al. \cite{li2017atten} propose a part-aligning CNN network for locating latent regions (hard attention) and then extract these regional features for Re-ID. Zhao et al. \cite{zhao2017atten} employ the Spatial Transformer Network \cite{stnj} as the hard attention model to find discriminative image parts. Li et al. \cite{attenli2018diversity}
use multiple spatial attention modules (by softmax function) to extract features at different spatial locations. Xu et al. \cite{xu2018attention} propose to mask the convolutional maps via pose-guided attention module. Li et al. \cite{hacnn} jointly learn multi-granularity attention selection and feature representation
for optimising person Re-ID in deep learning. However, most of the attention-based methods are often more prone to higher feature correlations, as these methods tend to have features focus on a more compact subspace, which makes the extracted features attentive but less diverse and therefore leads to suboptimal matching performance.

\subsection{Pose-driven Re-ID model}
While some pose-driven methods utilize pose information to tackle the occlusion or body-part missing problems. In these methods, pose landmarks are introduced to help to align body parts as pose landmarks indicate the body position of persons. Zheng et al. \cite{PBF} propose to use a CNN-based external pose estimator to normalize person images based on their pose and the
original and normalized images are then used to train a single deep Re-ID embedding. Sarfraz et al. \cite{pse} directly concatenate fourteen landmarks confidence maps with the image as network input, letting the model learns alignment in an automatic way. Huang et al. \cite{huang2018eanet} propose a Part Aligned Pooling (PAP) that utilizes seventeen human landmarks to enhance alignment. Miao et al. \cite{PGFA} learn to exploit pose landmarks to disentangle the useful information from the occlusion noise. However, the landmarks of persons are obtained usually by a third pose estimation model that trained on extra dataset, which increases the complicity of the whole Re-ID network. What’s more, standard pose estimation datasets may not cover the drastic viewpoint variations in surveillance scenarios, besides, surveillance images may not have sufficient resolution for stable landmarks prediction.

\begin{figure*}
\begin{center}
\includegraphics[width=0.9\linewidth]{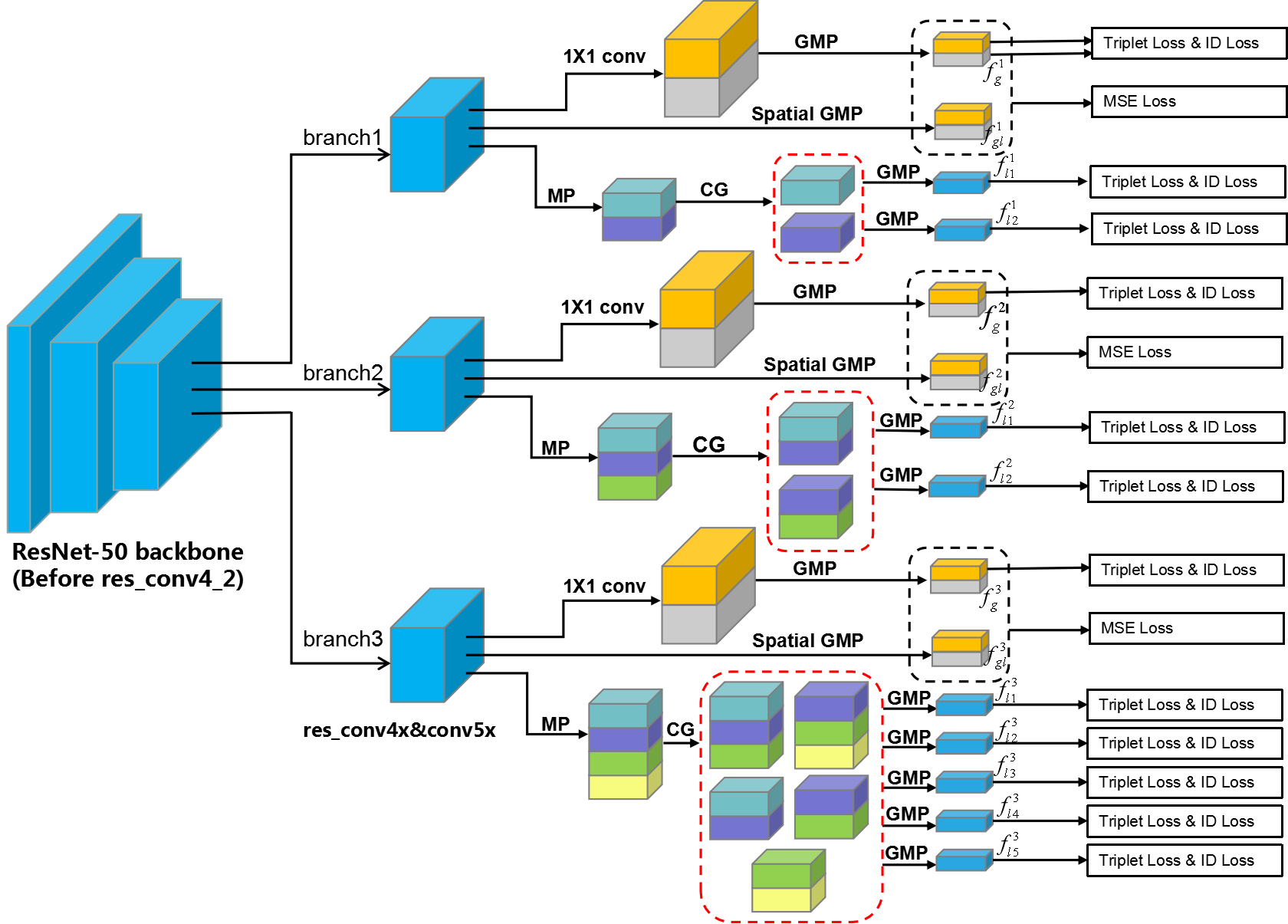}
\end{center}
	\caption{The structure of CGPN. The ResNet-50 backbone is split into three branches after \(res\_conv4\_1\) block. Each branch consists of a global part and a local part. In the global part, we apply two \(1 \times 1\) convolutional layers and global max pooling (GMP) to generate global features. While in the local part, we apply a max pooling(MP) with different kernel size, split the feature maps to different spatial horizontal stripes, then apply a coarse grained (CG) strategy and global max pooling (GMP) to generate local features.}
\label{fig:network}
\end{figure*}

\section{Proposed Method}
\subsection{Structure of CGPN}

In this part, we present our CGPN structure as in Fig. \ref{fig:network}. The backbone of our network is a CNN structure, such as ResNet \cite{he2016deep}, which achieves competitive results in many deep learning tasks. Like MGN\cite{mgn}, we divide the output of \(res\_conv4\_1\) into three different branches. Through the backbone network, CGPN transfers the input image into a 3D tensor \(T\) with size of \(c\times h\times w\) (\(c\) is channel number,
\(h\) is height, \(w\) is width). Each of the three branches contains a global part and a local part, the global part in three branches share the same structure while in every local part the output feature maps are uniformly partitioned into different strips.

In the global part, two \(1\times 1\) convolution layers are applied to the output feature maps to extract regional features. Each of the \(1\times 1\) convolution layers will output \( c\)-channel features and be supervised by the corresponding part features. In more detail, for \(i-\)th branch's global part, to supervise the global features, the output feature maps are uniformly divided into two parts in the vertical direction, and a global pooling is applied to each of them to extract two part features \{\(f_{gl1}^i\), \(f_{gl2}^i\)\}. The two part features \{\(f_{gl1}^i\), \(f_{gl2}^i\)\} are utilized in training stage to supervise global features \{\(f^i_{g1}\),\(f^i_{g2}\)\} generated by
the two \(1\times 1\) convolution layers. After training stage finishes, the two part features are no longer needed. The first \( c\)-channel global features \(f^i_{g1}\) should be closer to the upper part features \(f_{gl1}^i\) and in the same way, the second \(c\)-channel global features \(f^i_{g2}\) should be closer to the bottom part features \(f_{gl2}^i\). In inference stage, the first
\(c\)-channel global features \(f^i_{g1}\) and the second \(c\)-channel global features \(f^i_{g2}\) are concatenated to form \(2c\)-channel features as final global features \(f^i_g\), as the three branches' global part all share the same structure, we can get three global features \{\(f^1_g\), \(f^2_g\), \(f^3_g\)\} in total. With the supervision of the part features, in final \(2c\)-channel
global features, the first \(c\)-channel global features are forced to focus on the upper part of human body, while the second \(c\)-channel global features focus on the bottom part of human body, which makes final global features more robust to person image occlusion or body-part missing.

\begin{figure}
\begin{center}
\includegraphics[width=0.8\linewidth]{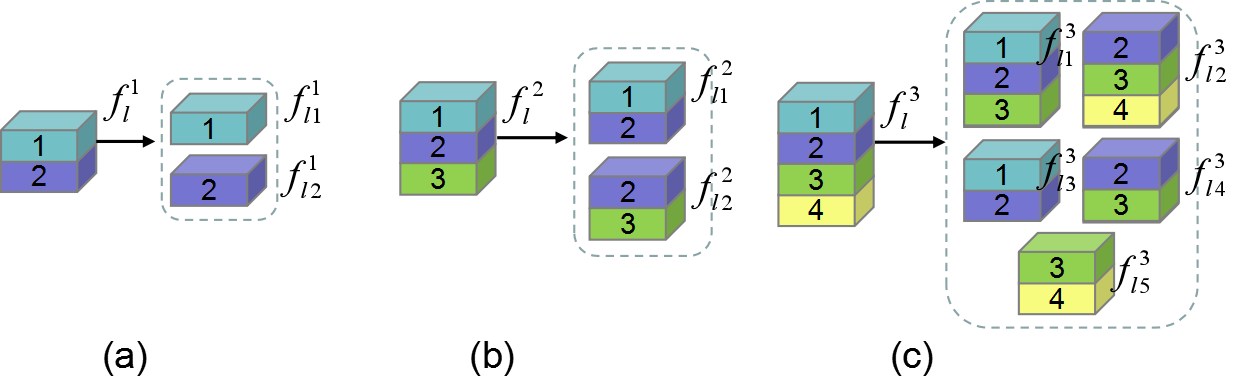}
\end{center}
\caption{The coarse grained part-level feature strategy. We keep only local features with height proportion no less than half of the whole feature maps' height. In local part of the first branch, the output feature maps are divided into two strips as (a) and we get two local features \{\(f^1_{l1}\), \(f^1_{l2}\)\} corresponding to size of \(c \times (h/2)\times w\). In local part of the second branch, the output feature maps are divided into three strips as (b), then we combine these strips and generate two local features \{\(f^2_{l1}\), \(f^2_{l2}\)\} corresponding to size of \(c\times (2h/3)\times w\). In local part of the third branch, the output feature maps are divided
into four strips as (c), then we combine these strips and generate five local features \{\(f^3_{l1}\), \(f^3_{l2}\), \(f^3_{l3}\), \(f^3_{l4}\), \(f^3_{l5}\)\} corresponding to size of \(c\times (3h/4)\times w\), \(c\times(3h/4)\times w\), \(c\times (h/2)\times w\), \(c\times (h/2)\times w\),
\(c\times (h/2)\times w\) respectively.}
\label{fig:coarse_grained_structure}
\end{figure}

For the local part in three branches, the output feature maps are divided into \(N\) stripes in the vertical direction with each strip having size of \(c \times (h/ N) \times w\), on which we prepare to extract local features. However, for person images that are occluded or body part missed, too much finer grained local features might be harmful and decrease the performance of
person Re-ID. To alleviate the drawbacks of fine grained local features, we choose to extract local features in a bigger receptive field which contains enough body structure information to well represent the corresponding body region. In this paper, we propose a coarse grained part-level
feature strategy in which the combined part strips must be adjacent and the minimum height proportion of combined local features should be no less than half of the output feature maps. The detail of the coarse grained strategy is illustrated in Fig. \ref{fig:coarse_grained_structure}, in local part of the first branch, the output feature maps are divided into two stripes in vertical
direction, then pooling operations are performed to get local feature representations \{\(f^1_{l1}\), \(f^1_{l2}\)\}. In the same way, for local part in the second branch, the output feature maps are divided into three stripes but we combine two adjacent strips to get two \(2/3\) proportion local features \{\(f^2_{l1}\), \(f^2_{l2}\)\}. For local part of the third branch, the output feature maps are divided into four stripes, then we combine two and three adjacent strips to get three \(1/2\) proportion and two \(2/3\) proportion local features respectively \{ \(f^3_{l1}\), \(f^3_{l2}\), \(f^3_{l3}\), \(f^3_{l4}\), \(f^3_{l5}\)\}.

During testing, all features \{\(f^1_g\), \(f^2_g\), \(f^3_g\), \(f^1_{l1}\), \(f^1_{l2}\), \(f^2_{l1}\), \(f^2_{l2}\), \(f^3_{l1}\), \(f^3_{l2}\), \(f^3_{l3}\), \(f^3_{l4}\), \(f^3_{l5}\)\} generated by global part the local part in each branch are reduced to 256-dim and are concatenated together as the final features, as different branches in CGPN actually learn representing information with different granularities which can cooperatively supplement discriminating information after the concatenation operation.

\subsection{Loss Functions}

Like various deep Re-ID methods, we employ softmax loss for classification, and triplet loss \cite{tripletloss} for metric learning. For the supervision of global part in each branch, we use mean square error loss (MSE) in training stage.
To be precise, in each branch, the local part is trained with the combination of softmax loss and triplet loss while the global part is trained with MSE loss, softmax loss and triplet loss as illustrated in Fig. \ref{fig:network}.

For \(i-\)th learned features \(f_i\), \(W_k\) is a weight vector for class \(k\) with the total calss number \(C\), \(N\) is the mini-batch in training stage, the softmax loss is formulated as,
\begin{equation}
	L_{softmax}=- \sum_{i=1}^{N}\log \frac{e^{W_{y_i}^{T}f_{i}}}{\sum_{k=1}^{C}{e^{W_{k}^{T}f_i}}}
\end{equation}

We employ the softmax loss to all global features \{\(f^1_g\), \(f^2_g\), \(f^3_g\)\} and all coarse grained local features \{\({f_{li}^1|_{i=1}^2}\), \({f_{li}^2|_{i=1}^2}\), \({f_{li}^3|_{i=1}^5}\)\}.
Besides, all the global features are also trained with triplet loss. In training stage, an improved batch hard triplet loss is applied with formula as follows,
\begin{equation}
\begin{aligned}
L_{triplet}=\sum_{i=1}^{P}\sum_{a=1}^{K} [
\alpha +\mathop{max}\limits_{p=1...K}\left \| f_a^{(i)}-f_p^{(i)} \right \|_2 - \\
\mathop{min}\limits_{\substack{n=1...K\\j=1...P\\j\neq i}}\left \| f_a^{(i)} - f_n^{(j)} \right \|_2
]_+
\end{aligned}
\end{equation}

In above formula, P is the number of selected identities and K is the number of images from each identity in a mini-batch. \(f_a^{(i)}\) is anchor sample, \(f_p^{(i)}\)is positive sample, \(f_n^{(i)}\) is negative sample and \(\alpha\) is the margin for training triplet, it is set to \(1.2\) in our implementation.

To supervise the global features, we employ the MSE loss to all global features \{\(f^1_g\), \(f^2_g\), \(f^3_g\)\} and supervision local features \{\(f^1_{gl}\), \(f^2_{gl}\), \(f^3_{gl}\)\} with formula as follows,
\begin{equation}
L_{mse} = \sum_{i=1}^B\sum_{p=1}^M\|f_{gp}^i - f_{glp}^i\|_2^2
\end{equation}

Where \(f_{gp}^i\) is the \(p-\)th \(c\)-channel feature map of global features in \(i-\)th branch, \(f_{glp}^i\) is the corresponding \(p-\)th \(c\)-channel local part in \(i-\)th branch, as the global part consists of two \(c\)-channel features and there are three branches in our network, M is set to 2 and B is set to 3 in our implementation.

The overall training loss is the sum of above three loss, which is formulated by,
\begin{equation}
L = L_{softmax} + L_{triplet} + L_{mse}
\end{equation}

\section{Experiment}

\subsection{Datasets and Protocols}

We train and test our model respectively on three mainstream Re-ID datasets: Market1501 \cite{market1501}, DukeMTMC-reID \cite{dukemtmc}, CUHK03 \cite{cuhk03}. Especially the CUHK03 dataset, it is the most challenging realistic scenario Re-ID dataset as it consists a lot of occluded or body part missed pedestrian images as illustrated in Fig. \ref{fig:cuhk03_data}.

Market-1501 is captured by six cameras in front of a campus supermarket, which contains 1501 person identities, 12936 training images from 751 identities and 19732 testing images from 750 identities. All provided pedestrian bounding boxes are detected by DPM \cite{DPM}.

DukeMTMC-reID contains 1812 person identities captured by 8 high-resolution cameras. There are 1404 identities appear in more than two cameras and the other 408 identities are regarded as distractors. The training set consists of 16552 images from 702 identities and the testing set contains 17661 images from the rest 702 identities.

CUHK03 contains 1467 person identities captured by six cameras in CUHK campus. Both manually labeled pedestrian bounding boxes and automatically detected bounding boxes are provided. In this paper, we use the manually labeled version and we follow the new training/testing protocol proposed in \cite{protocolcuhk03}, 7368 images from 767 identities for training and 5328 images from 700 identities for testing.

In our experiment, we report the average Cumulative Match Characteristic (CMC) at Rank-1 and mean Average Precision (mAP) on all the candidate datasets to evaluate our method.

\subsection{Implementation Details}

All images are re-sized into \(384 \times 128\) and the backbone network is ResNet-50 \cite{he2016deep} pre-trained on ImageNet with original fully connected layer discarded. In training stage, the mini-batch size is set to 64, in which we randomly select 8 identities and 8 images for each identity (\(P=8, K=8\)). Besides, we deploy a randomly horizontal flipping strategy to images for data augmentation. Different branches in the network are all initialized with
the same pre-trained weights of corresponding layers after \(res\_conv4\_1\) block. Our model is implemented on Pytorch platform. We use SGD as the optimizer with the default hyper-parameters (momentum=0.9, weight decay factor=0.0005) to minimize the network loss. The initial learning rate is set to 1e-2 and we decay it at epoch 60 and 80 to 1e-3 and 1e-4 respectively. The total training takes 240 epochs. During evaluation, we use the average of original images features and horizontally flipped images features as the final features. All our experiments on different datasets follow the settings above.

\subsection{Comparison with State-of-the-Art Methods}

In this section, we compare our proposed approach with current state-of-the-art methods on the three main-stream Re-ID datasets.
\begin{table*}
    \begin{center}

    \setlength{\tabcolsep}{2mm}
        \begin{tabular}{c|cc|cc|cc}
            \hline
            \multicolumn{1}{c|}{\multirow{2}{*}{Method}} & \multicolumn{2}{c|}{Market1501} & \multicolumn{2}{c|}{DukeMTMC-reID} &\multicolumn{2}{c}{CUHK03} \\ \cline{2-7} \multicolumn{1}{c|}{}&Rank-1&mAP &Rank-1&mAP&Rank-1&mAP \\

            \hline\hline
            IDE \cite{ide} &-&-&-&-& 22.2 & 21.0 \\
            PAN \cite{pan} &-&-&-&-& 36.9 & 35.0\\
            SVDNet \cite{svdnet} &-&-&-&-& 40.9 & 37.8\\
            MGCAM \cite{mgcam} & 83.8 & 74.3 &-&-& 50.1 & 50.2\\
            HA-CNN \cite{hacnn} & 91.2 & 75.7& 80.5 & 63.8& 44.4 & 41.0 \\
            VPM \cite{vpm} & 93.0 & 80.8 & 83.6 & 72.6 &-&- \\
            SCP \cite{scpnet} & 94.1 & - &84.8 & - &-&- \\
            PCB+RPP \cite{pcb} & 93.8 & 81.6& 83.3 & 69.2&-&- \\
            SphereReID \cite{spherereid} & 94.4 & 83.6 & 83.9 & 68.5&-&- \\
            MGN \cite{mgn} & 95.7 & 86.9& 88.7 & 78.4& 68.0 & 67.4 \\
            DSA \cite{DSA} & 95.7 & 87.6& 86.2 & 74.3& 78.9 & 75.2 \\
            Pyramid \cite{pyramid} & 95.7 & 88.2& 89.0 & 79.0& 78.9 & 76.9 \\
            SAN \cite{san} & 96.1 & 88.0& 87.9 & 75.5& 80.1 & 76.4 \\
            \hline
            CGPN & $\mathbf{96.1}$ & $\mathbf{89.9}$& $\mathbf{90.4}$ &$\mathbf{80.9}$& $\mathbf{87.1}$ & $\mathbf{83.6}$\\
            \hline
        \end{tabular}
    \end{center}
    \caption{Performance (\%) comparisons to the state-of-the-art results on Market1501,DukeMTMC-reID and CUHK03 datasets at evaluation metric of Rank-1 and mAP in single query mode without re-ranking.}
\label{tab:dataset}
\end{table*}


The statistical comparison between our CGPN Network and the state-of-the-art methods on Market-1501, DukeMTMCreID and CUHK03 datasets is shown in Table \ref{tab:dataset}. On Market-1501 dataset, SAN achieved the best published result without re-ranking , but our CGPN achieves 89.9\% on metric mAP, exceeding SAN by +1.9\%. On metric Rank-1, our CGPN achieves 96.1\%, on par with SAN while our model is trained in an easier and end to end way.
Compared with MGN which is also a multiple branches method, our model surpasses MGN by +0.4\% on metric Rank-1 and by +3.0\% on metric mAP.
Among the compared results on DukeMTMC-reID dataset, Pyramid achieved the best published result on metric Rank-1 and mAP respectively. Our CGPN achieves state-of-the-art result of Rank-1/mAP = 90.4\%/80.9\%, outperforming Pyramid by +1.4\% on metric Rank-1 and +1.9\% on metric mAP.
From Table \ref{tab:dataset}, our CGPN model achieves Rank-1/mAP = 87.1\%/83.6\% on the most challenging CUHK03 labeled dataset under the new protocol. On metric Rank-1, our CGPN outperforms the best published result of SAN by +7.0\% and outperforms the best published result of Pyramid by +6.7\% on mAP.

In summary, our proposed CGPN can always outperform all other existing methods and shows strong robustness over different Re-ID datasets. According to the comparative experiments on the three datasets, especially on CUHK03 dataset, our approach can consistently outperform all other models by a large margin and is the first method to improve the performance on CUHK03 to above 80\% on both metric Rank-1 and mAP under the new protocol without re-ranking. Therefore, we can conclude that our method can effectively extract robust deep features for
occluded or body part missed pedestrian images in person Re-ID.

\begin{table*}
\begin{center}

\setlength{\tabcolsep}{2mm}
\begin{tabular}{c|c|c|c|c|c|c}
\hline
\multirow{2}{*}{Method} &
\multicolumn{2}{c|}{Market1501}&
\multicolumn{2}{c|}{DukeMTMC-reID}&
\multicolumn{2}{c}{CUHK03}\\
\cline{2-7}
&Rank-1&mAP&Rank-1&mAP&Rank-1&mAP\\
\hline\hline
CGPN-1 & 94.9 & 87.9 & 89.3 & 78.4 & 82.4 & 79.9\\
CGPN-2 & 95.3 & 89.4 & 90.3 & 80.2 & 85.3 & 82.5\\
CGPN-3 & 94.2 & 86.2 & 88.6 & 76.9 & 84.3 & 81.0\\
CGPN-4 & 95.2 & 89.3 & 90.0 & 79.9 & 83.4 & 80.7\\
CGPN & $\mathbf{96.1}$ & $\mathbf{89.9}$ & $\mathbf{90.4}$ &
$\mathbf{80.9}$ & $\mathbf{87.1}$ & $\mathbf{83.6}$\\
\hline
\end{tabular}
\end{center}
\caption{Ablation study of CGPN coarse grained part-level feature strategy and
supervised global part. Comparison results(\%) on Market1501, DukeMTMC-reID and
CUHK03-labeled at evaluation metric of Rank-1 and mAP in single query mode without re-ranking.}
\label{tab:ablation_clp}
\end{table*}

\subsection{The importance of coarse grained part-level features}

\begin{figure}[t]
	\begin{center}
		\includegraphics[width=0.8\linewidth]{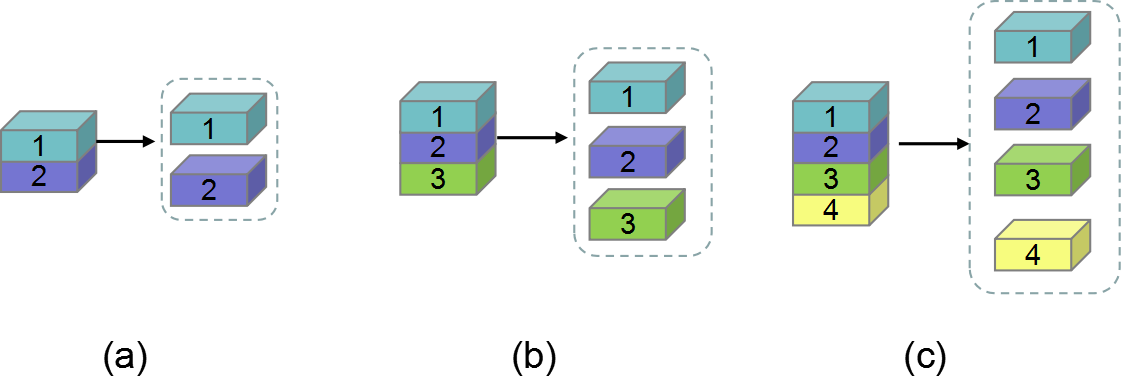}
	\end{center}
	\caption{Fine grained local part structure in CGPN-2. The output feature map in each branch is uniformly split into two, three, four stripes in vertical direction respectively.}
	\label{fig:fine_grained_strategy}
\end{figure}

To verify the effectiveness of coarse grained part-level feature strategy in CGPN model, We train two mal-functioned CGPN for comparison:
\begin{itemize}
	\item CGPN-1 abandons the local part in three branches and keeps only global parts.
	\item CGPN-2 replaces coarse grained part-level features with fine grained part-level features. It abandons coarse grained strategy in local part of three branches, compared with normal CGPN model, its local part in three branches directly divide output feature map into two, three and four parts as in Fig. \ref{fig:fine_grained_strategy}.
\end{itemize}

Comparing CGPN-1 with CGPN, we can see a significantly performance decrease on
Rank-1/mAP by -1.2\%/-2.0\%, -1.1\%/-2.5\% and -4.7\%/-3.7\% on Market1501,
DukeMTMC-reID and CUHK03 dataset respectively. Especially on CUHK03, we can observe a sharp decrease by -4.7\%/-3.7\% on metric Rank-1/mAP. As CGPN-1 is trained in exactly the same procedure with CGPN model and CUHK03 dataset typically consists of many occluded or body part missed person images, we can infer that the coarse grained local part is critical for CGPN model, especially on the dataset which contains a lot of occluded or body part missed
person images.

Comparing CGPN-2 with CGPN, we can still observe a performance decrease by -0.8\%/-0.5\%, -0.1\%/-0.7\% and -1.8\%/-1.1\% on metric Rank-1/mAP on Market1501, DukeMTMC-reID and CUHK03 datasets respectively. Compared with fine grained part-level features, coarse grained part-level features contain enough body structure information to better represent the corresponding
body regions which make CGPN learn more robust local features. Besides, on CUHK03, we can also see a sharper performance decrease compared with the other two datasets. The reason is that Market1501 and DukeMTMC-reID consist of mainly holistic person images with little occlusions or body part missing, these images keep complete body structure and make fine grained part-level features achieve comparable performance with coarse grained part-level features. While on
CUHK03, as it consists of a lot of occluded or body part missed person images, coarse grained part-level features outperform fine grained part-level features evidently. Our experiments clearly prove that our coarse grained part-level feature strategy can improve model performance significantly and is critical for model robustness, especially for occluded or body part missed person images.

\subsection{The importance of supervised global part}

To further verify the effectiveness of supervised global part in CGPN model, We train another two mal-functioned CGPN for comparison:
\begin{itemize}
\item CGPN-3 abandons global parts in all three branches and keeps only local parts which are trained with triplet loss and softmax loss.
\item CGPN-4 keeps the global parts but abandons the supervision learning of all global parts in three branches, these global parts are trained only with triplet loss and softmax loss.
\end{itemize}

Comparing CGPN-3 with CGPN, we observe a dramatic performance decrease on all three datasets, the performance on metric Rank-1/mAP decrease by -1.9\%/-3.7\%, -1.8\%/-4.0\% and -2.8\%/-2.6\% on Market1501, DukeMTMC-reID and CUHK03 respectively. As three models are trained in exactly the same procedure, we conclude that the global part is critical to CGPN.
Comparing CGPN-4 with CGPN, after abandoning global supervision, we observe a performance decrease on Rank-1/mAP of -0.9\%/-0.6\% and -0.4\%/-1.0\% on Market1501 and DukeMTMC-reID. While on CUHK03 we observe a dramatic performance decrease by -3.7\%/-2.9\%. The reason of such a different performance decrease is that Market1501 and DukeMTMC-reID mainly consist of holistic person images with which the global part can get enough good global features directly even without supervision, while CUHK03 contains a lot of occluded or body part missed person images and the supervised global part is much more important for extracting accurate global features.

Comparing CGPN-4 with CGPN-3, after adding unsupervised global parts, we see a large performance improvement on Rank-1/mAP of +1.0\%/+3.1\% and +1.4\%/+3.0\% on Market1501, DukeMTMC-reID. But on CUHK03 we observe a significant performance decrease by -0.9\%/-0.3\% unexpectedly. As analyzed above, the image type is quite different for the three datasets, especially CUHK03 contains a lot of occluded or body part missed person images, the unexpected performance decrease on CUHK03 further proves that unsupervised global features can be harmful and certainly impair model performance. We conclude that the supervision of global features is critical for high performance of person Re-ID and that unsupervised global features will result in inaccurate global features which impair model performance evidently.

\begin{figure*}
	\begin{center}
		\includegraphics[width=0.86\linewidth,height=0.65\textheight]{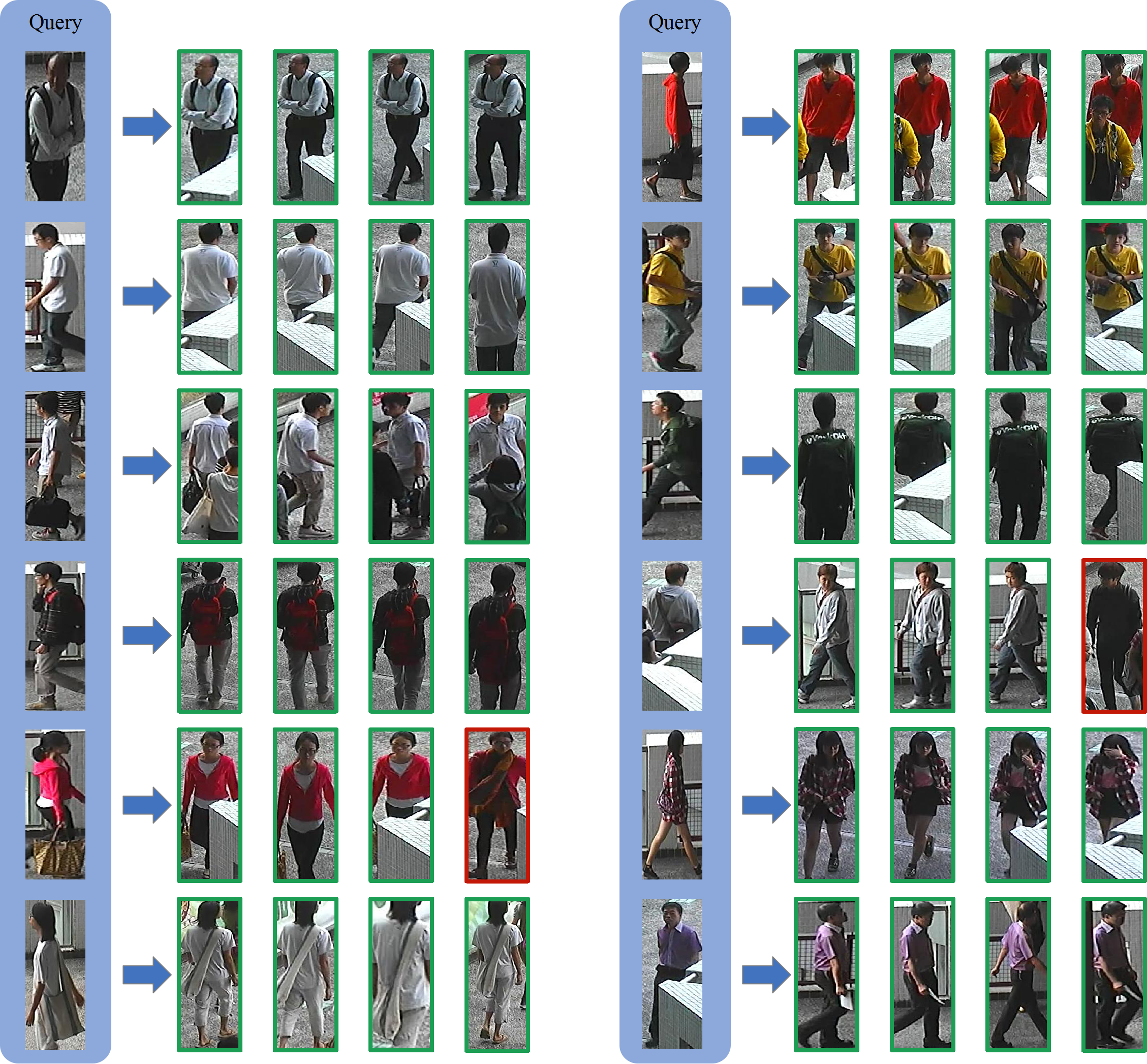}
	\end{center}
	\caption{Top-4 ranking list for some query images on CUHK03-labeled dataset by CGPN. The retrieved images are all from
		the gallery set, but not from the same camera shot. The images with green borders belong to the same identity as the given
		query, and that with red borders do not.}
	\label{fig:cuhk03_retrieval_result}
\end{figure*}


\subsection{Visualization of Re-ID results}

We visualize the retrieval results by CGPN and MGN for some given query pedestrian images of CUHK03-labeled dataset in Fig.\ref{fig:cuhk03_retrieval_result}. These retrieval results show the
great robustness of our CGPN model, regardless of the occlusions or body part missing of detected pedestrian images, CGPN can robustly extract discriminative features for different identities.

\section{Conclusion}

In this paper, we propose a coarse grained part-level features learning network (CGPN) integrated with supervised global-level features for person Re-ID. With the coarse grained part-level strategy, the local parts in three branches learn more discriminative local features. With the supervision learning of global parts in three branches, the global parts learn to extract more accurate and
suitable global features for pedestrian images. Experiments have confirmed that our model not only achieves state-of-the-art results on all three main-stream person Re-ID datasets, but also pushes the performance to an exceptional level.

{\small
\bibliographystyle{ieee_fullname}
\bibliography{egbib}
}

\end{document}